\documentclass[conference]{IEEEtran} 
\IEEEoverridecommandlockouts 

\usepackage{cite}                  
\usepackage{amsmath,amssymb,amsfonts}
\usepackage{graphicx}
\usepackage{array}
\usepackage{booktabs}
\usepackage{tabularx}
\usepackage{url}
\usepackage{hyperref}
\hypersetup{hidelinks}             
\usepackage{xcolor}

\usepackage{array,tabularx,booktabs,ragged2e}
\newcolumntype{P}[1]{>{\RaggedRight\arraybackslash}p{#1}}
\newcolumntype{W}[1]{>{\RaggedRight\arraybackslash\hsize=#1\hsize}X}

\title{From Autoencoders to CycleGAN: Robust Unpaired Face Manipulation via Adversarial Learning}

\author{
\IEEEauthorblockN{Collin Guo}
\IEEEauthorblockA{College of Information\\
University of Maryland - College Park\\
\texttt{cyguo@terpmail.umd.edu}}
\and
\IEEEauthorblockN{Yi Qian}
\IEEEauthorblockA{Department of Computer Science\\
The University of Tulsa\\
\texttt{yi-qian@utulsa.edu}}
}

\begin{document}
\maketitle
\begin{abstract}
Human face synthesis and manipulation are increasingly important in entertainment and AI, with a growing demand for highly realistic, identity-preserving images even when only unpaired, unaligned datasets are available. We study \emph{unpaired} face manipulation via adversarial learning, moving from autoencoder baselines to a robust, guided CycleGAN framework. While autoencoders capture coarse identity, they often miss fine details. Our approach integrates spectral normalization for stable training, identity- and perceptual-guided losses to preserve subject identity and high-level structure, and semantic/landmark-weighted cycle constraints to maintain facial geometry across pose and illumination changes. Experiments show that our adversarially trained CycleGAN improves realism (FID), perceptual quality (LPIPS), and identity preservation (ID-Sim) over autoencoders, with competitive \emph{cycle-reconstruction} PSNR/SSIM and practical inference times—achieving high quality without paired datasets and approaching pix2pix on curated paired subsets. These results demonstrate that guided, spectrally normalized CycleGANs provide a practical path from autoencoders to \emph{robust unpaired} face manipulation.
\end{abstract}

\section{Introduction}	
In recent years, deepfake applications have brought face-changing technologies into the mainstream, demonstrating the ability of generative models to produce highly realistic human faces. Facial manipulation tasks generally fall into three main categories: face synthesis, face swapping, and facial attribute/expression editing. From a technical standpoint, these applications are built upon advances in deep generative modeling, including autoencoders, generative adversarial networks (GANs), and adversarial architectures for high-quality image-to-image translation such as pix2pix (paired) and CycleGAN (unpaired) \cite{pix2pix,zhu2017cyclegan}.

Despite impressive progress, many existing approaches remain limited to demo-style implementations rather than scalable, production-ready solutions. Real-world applications must overcome key challenges: efficiently processing large-scale datasets; stabilizing adversarial training to prevent overfitting, mode collapse, or discriminator blow-ups; and achieving fine-grained, identity-faithful attribute transfer under pose, illumination, and domain shifts. Autoencoder-based methods capture broad facial structure but often blur high-frequency details; paired translation excels with curated supervision yet fails to generalize without alignment; and vanilla unpaired translation can drift in identity and lose local texture.

This project addresses these limitations by moving from autoencoder-based facial manipulation to an improved CycleGAN framework that leverages adversarial learning for more robust and flexible results. Concretely, we incorporate spectral normalization to constrain the discriminator’s Lipschitz behavior and stabilize training \cite{spectralnorm}; identity- and perceptual-guided learning (frozen face encoder and VGG-Face-style features) to preserve subject identity and high-level structure \cite{larsen2016aebp}; semantic/landmark-weighted cycle consistency to maintain facial geometry; and lightweight architectural refinements—patchwise contrastive regularization for local detail, multi-scale discriminators, and optional attention in the generator—to reduce artifacts and improve coherence \cite{karras2019stylegan}. We further adopt practical training refinements (TTUR, EMA, and differentiable augmentations) that improve convergence smoothness and sample efficiency.

Our work builds upon a baseline deepfake-style autoencoder implementation \cite{deepfake} for learning facial representations and extends it with CycleGAN \cite{zhu2017cyclegan} to produce clearer, more realistic, and adaptable manipulations. Evaluated on unpaired frames from a public debate video and curated VGGFace2 subsets (with an additional human$\leftrightarrow$anime stress test), the proposed system improves realism (FID), perceptual quality (LPIPS), and identity preservation (ID-Sim) over autoencoders, while approaching pix2pix on curated paired subsets—without requiring paired training data. These enhancements lead to faster, more stable convergence and reduced reliance on pairing, providing a pathway toward practical, scalable face manipulation systems.

\section{Related Work}
\begin{figure*}[!t]
    \centering
    \includegraphics[width=11cm]{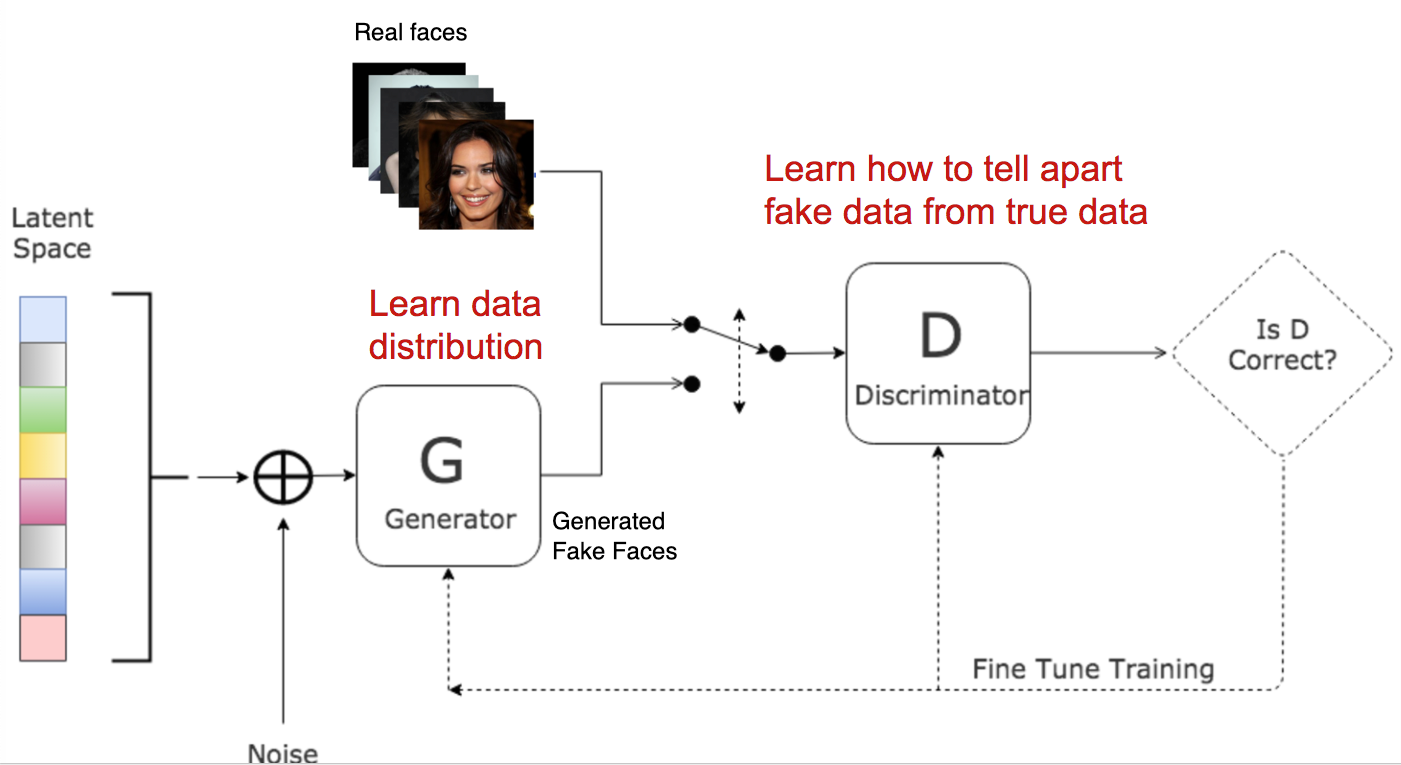}
    \caption{GAN architecture and compute process.}
    \label{fig:gan-arch}
\end{figure*}

Deep generative modeling has progressed along three principal lines:
\emph{autoregressive models (AR)}, \emph{variational autoencoders (VAEs)}, and
\emph{generative adversarial networks (GANs)}.
Autoregressive approaches factorize the joint density into a product of conditional
distributions, which yields tractable likelihoods but can be slow at sample time due to
their sequential nature \cite{ostrovski2018autoregressive}. VAEs introduce a probabilistic latent space with amortized inference \cite{Kingma2014,kingma2014vae}.
Given $x$, the encoder produces parameters $(\mu(x), \sigma(x))$ of a Gaussian
approximate posterior $q_{\phi}(z\!\mid\!x)$, and reparameterization draws
\begin{equation}
  z = \mu(x) + \sigma(x)\odot \epsilon, \quad \epsilon \sim \mathcal{N}(0,I).
\end{equation}
The decoder defines $p_{\theta}(x\!\mid\!z)$, and learning maximizes the evidence lower
bound (ELBO):
\begin{equation}
\begin{aligned}
  \mathcal{L}_{\text{VAE}}(x)
  &= \mathbb{E}_{q_{\phi}(z\mid x)}\!\left[-\log p_{\theta}(x\mid z)\right]\\
  &\quad  + \beta\, D_{\mathrm{KL}}\!\left(q_{\phi}(z\mid x)\,\|\,p(z)\right).
\end{aligned}
\end{equation}
Extensions improve perceptual fidelity by comparing in learned feature spaces
\cite{larsen2016aebp} and by blending variational inference with adversarial training
\cite{bao2017cvaegan}.

GANs formulate generation as a two-player minimax game between a generator $G$ and a
discriminator $D$ \cite{gan}. The generator maps latent noise
$z \sim p_{z}$ (e.g., standard Gaussian) to the data space, producing $G(z)$,
while the discriminator receives either real data $x \sim p_{\text{data}}$ or synthetic
samples $G(z)$ and outputs the probability of being real. The standard
value function is

\begin{equation}
\begin{aligned}
\min_{G}\max_{D}\, V(D,G)
&= \mathbb{E}_{x\sim p_{\text{data}}}\!\left[\log D(x)\right] \\
&\quad + \mathbb{E}_{z\sim p_{z}}\!\left[\log\!\bigl(1 - D(G(z))\bigr)\right].
\end{aligned}
\end{equation}

Training alternates gradient updates to $D$ and $G$; at equilibrium, the model distribution
$p_{g}$ matches $p_{\text{data}}$ and $D$ cannot discriminate better than chance.
A large body of subsequent work explores improved architectures and training objectives,
including conditional GANs for paired translation \cite{pix2pix},
feature-metric regularization for autoencoding \cite{larsen2016aebp},
hybrid variational--adversarial models such as CVAE-GAN \cite{bao2017cvaegan},
and spectral normalization for stabilizing training \cite{spectralnorm}.
State-of-the-art unconditional generators (e.g., StyleGAN) have further advanced image
quality via style-based modulation and progressive growing \cite{karras2019stylegan}.

CycleGAN extends adversarial learning to \emph{unpaired} image–to–image translation by
learning two mappings between domains $A$ and $B$—a forward generator $G\!:\!A\!\to\!B$
and a backward generator $F\!:\!B\!\to\!A$—together with discriminators $D_A$ and $D_B$
that judge whether translated images look like real samples from their respective domains
\cite{zhu2017cyclegan}. During training, the discriminators encourage each
generator to produce outputs that are indistinguishable from target-domain images
(adversarial learning), while a cycle-consistency constraint requires that translating to
the other domain and back approximately recovers the original input. This combination
preserves content and layout without paired supervision, which is especially useful in
face manipulation where pose, expression, and illumination differ across domains.
For practical use, widely adopted PyTorch implementations are available
\cite{cyclegan_repo,cyclegan_github}.

Recent diffusion models generate high-fidelity images by denoising in a learned latent
space, dramatically improving sample quality and controllability
\cite{rombach2022ldm}. Although diffusion is beyond the scope of our main method,
we include it for completeness as a contemporary alternative to GAN-based approaches.

\section{Dataset and Features}

\begin{figure*}[!t]
    \centering
    \includegraphics[width=10cm]{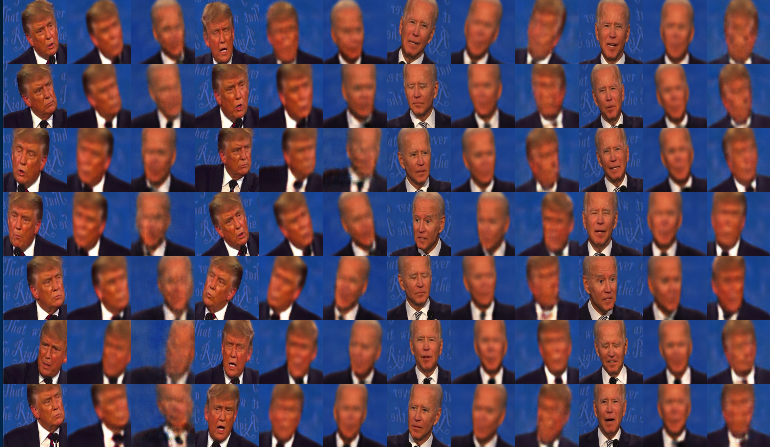}
    \caption{Deep face manipulation examples from the 2024 U.S.\ Presidential Debate.}
    \label{fig:deepface-2020}
\end{figure*}

We extracted $\sim$1{,}000 face frames from the 2024 U.S.\ Presidential Debate video using
FFmpeg \cite{ffmpeg} for model training and validation ($\sim$500 per identity).
We also utilize VGGFace2 \cite{vggface2} for large-scale pose/diversity coverage
and reference the CelebA attributes dataset \cite{liu2015celeba} as a standard
benchmark for identity/attribute supervision. In addition, we experimented with
domain-biased content using the Anime Face Dataset from Kaggle \cite{animeface_kaggle}
to assess cross-domain transferability. These datasets allow evaluating both within-domain latent operations (VAE) and cross-domain translation (CycleGAN). Figure~\ref{fig:deepface-2020} shows an
example of face manipulation via deepfake-style techniques \cite{deepfake}.

\section{Methods}
\label{sec:methods}

\subsection{Problem Setup and Baseline}
We consider unpaired translation between two facial domains $X$ and $Y$ with data distributions
$p_X$ and $p_Y$. Following CycleGAN \cite{zhu2017cyclegan}, we learn generators
$G_{X\to Y}$ and $G_{Y\to X}$ and discriminators $D_Y$ and $D_X$ trained adversarially \cite{gan}
with cycle-consistency. For $X\!\to\!Y$ the adversarial objective is
\begin{equation}
\begin{aligned}
\mathcal{L}_{\mathrm{GAN}}(G_{X\to Y},&D_Y) = \mathbb{E}_{y\sim p_Y}\!\big[\log D_Y(y)\big]\\
&+ \mathbb{E}_{x\sim p_X}\!\big[\log(1-D_Y(G_{X\to Y}(x)))\big].
\end{aligned}
\end{equation}
Cycle-consistency enforces approximate invertibility:
\begin{equation}
\begin{aligned}
\mathcal{L}_{\mathrm{cyc}}(G_{X\to Y},
&G_{Y\to X}) = \\
&\quad \mathbb{E}_{x\sim p_X}\!\big[\|G_{Y\to X}(G_{X\to Y}(x)) - x\|_1\big]\\
&+ \mathbb{E}_{y\sim p_Y}\!\big[\|G_{X\to Y}(G_{Y\to X}(y)) - y\|_1\big].
\end{aligned}
\end{equation}
As in \cite{zhu2017cyclegan}, we use ResNet-based generators (9 residual blocks),
PatchGAN discriminators, and optimize with Adam \cite{adam}. Our generators are image-to-image
mappings (they take an image from the other domain as input rather than white noise), which is standard for CycleGAN.

\subsection{Spectral Normalization and Lipschitz Control}
We regularize the generator and discriminator via \emph{spectral normalization} (SN) \cite{spectralnorm}.
The discriminator’s function class critically affects GAN stability; constraining it to be
(approximately) Lipschitz-continuous prevents gradient explosions and improves boundedness of statistics.
Under the original GAN objective, the optimal discriminator has the form
\begin{equation}
D^\star(x)=\frac{p_{\mathrm{data}}(x)}{p_{\mathrm{data}}(x)+p_G(x)}=\sigma(f(x)),
\end{equation}
where $\sigma$ denotes the logistic sigmoid.
Moreover, its score function satisfies
\begin{equation}
\nabla_x f(x) \;=\; \frac{1}{p_{\mathrm{data}}(x)} \nabla_x p_{\mathrm{data}}(x)
\;-\; \frac{1}{p_G(x)} \nabla_x p_G(x),
\end{equation}
so bounding $\|\nabla_x f(x)\|$ provides practical Lipschitz control. SN enforces a per-layer
Lipschitz bound by replacing each weight matrix $W$ with $W/\sigma(W)$, where $\sigma(W)$ is
the largest singular value, estimated efficiently by power iteration:
$v_{t+1}\!\leftarrow\! \frac{W^\top W v_t}{\|W^\top W v_t\|}$ and $\sigma(W)\!\approx\!\|Wv_t\|$.
We apply SN to \emph{all} layers of $D$ and optionally to $G$’s convolutions (noting that SN
on $G$ can slightly reduce capacity).

\subsection{Proposed Improvements for Face Manipulation}
To make CycleGAN practical for identity-preserving, high-fidelity face manipulation, we introduce the following components.

\paragraph{Identity- and Perceptual-Guided Translation.}
Let $E(\cdot)$ be a frozen face-embedding network. We add an identity loss to preserve subject identity:
\begin{equation}
\begin{aligned}
\mathcal{L}_{\mathrm{id}}
& = \big(1 - \cos( E(G_{X\to Y}(x)), E(x) )\big)\\
& + \big(1 - \cos( E(G_{Y\to X}(y)), E(y) )\big).
\end{aligned}
\end{equation}
Beyond pixels, we include a feature/perceptual loss \cite{larsen2016aebp} using a frozen backbone $\Phi$
(e.g., VGG-Face features):
\begin{equation}
\begin{aligned}
\mathcal{L}_{\mathrm{perc}}
&= \sum_{\ell}\!\big\|\Phi_\ell(G_{X\to Y}(x)) - \Phi_\ell(y^\star)\big\|_1\\
&+ \sum_{\ell}\!\big\|\Phi_\ell(G_{Y\to X}(y)) - \Phi_\ell(x^\star)\big\|_1
\end{aligned}
\end{equation}
where $(x^\star,y^\star)$ are “pseudo-pairs” retrieved as nearest neighbors across domains in $\Phi$-space
from the current minibatches $\mathcal{B}_X,\mathcal{B}_Y$ (no gradient flows through the retrieval).

\paragraph{Semantic- and Landmark-Consistent Cycles.}
Let $S(\cdot)\!\in\![0,1]^{H\times W}$ be a precomputed face-parsing mask (skin, hair, lips, eyes).
We reweight cycle-consistency to prioritize facial regions:
\begin{equation}
\begin{aligned}
\mathcal{L}_{\mathrm{sem\text{-}cyc}}
&= \mathbb{E}_{x}\!\big[\| S(x)\!\odot\!(G_{Y\to X}(G_{X\to Y}(x)) - x)\|_1\big]\\
&+ \mathbb{E}_{y}\!\big[\| S(y)\!\odot\!(G_{X\to Y}(G_{Y\to X}(y)) - y)\|_1\big],
\end{aligned}
\end{equation}
with $S(\cdot)$ treated as constant (no gradients). With $K$ facial landmarks $L(\cdot)\!\in\!\mathbb{R}^{K\times 2}$
(precomputed and detached) we impose geometric consistency:
\begin{equation}
\begin{aligned}
\mathcal{L}_{\mathrm{lmk}}
& = \| L(G_{X\to Y}(x)) - L(x)\|_2 \\
&+ \| L(G_{Y\to X}(y)) - L(y)\|_2.
\end{aligned}
\end{equation}

\paragraph{Patchwise Contrastive Structure (Unpaired).}
To preserve local detail without paired supervision, we sample $N$ spatial patches from intermediate features of
$x$ and $G_{X\to Y}(x)$ and apply a patch-level contrastive loss:
\begin{equation}
\mathcal{L}_{\mathrm{con}}
= -\frac{1}{N}\sum_{i=1}^{N}
\log \frac{\exp(\mathrm{sim}(f(p_i^x), f(p_i^{Gx}))/\tau)}
{\sum_{j}\exp(\mathrm{sim}(f(p_i^x), f(p_j^{Gx}))/\tau)},
\end{equation}
where $\mathrm{sim}(u,v)$ is cosine similarity, $f$ is a projection head, positives are same-location patches,
and negatives are other locations within the same feature map and batch.

\paragraph{Attention and Multi-Scale Discrimination.}
We insert lightweight self-attention at $1/16$ and $1/8$ resolutions in $G$ to capture long-range dependencies
(hairlines, jaw contours) and adopt \emph{multi-scale} discriminators $D^1,D^2$ (full and half resolution)
to detect artifacts at different scales. Inspired by style-based generators \cite{karras2019stylegan}, we add
channel-wise modulation (AdaIN) in late residual blocks to stabilize style transfer.

\paragraph{Stabilization and Efficiency.}
Alongside SN \cite{spectralnorm}, we use global-norm gradient clipping, an exponential moving average (EMA) of
generator weights, and the Two-Time-Scale Update Rule (TTUR): two $D$ updates per $G$ update early in training.
We apply differentiable data augmentation (translation, color jitter, cutout) to $D$’s inputs, use mixed-precision
training (FP16), and progressively resize inputs ($128\!\to\!256$). For completeness, latent diffusion
\cite{rombach2022ldm} provides strong generative priors, but we keep our method fully GAN-based for efficiency.
Also, for hybrid supervisonion that when a small paired subset is available (e.g., pose-aligned frames), we add a conditional loss in the spirit of
pix2pix \cite{pix2pix} for those minibatches, while retaining the unpaired objectives elsewhere.

Our final loss is a weighted sum:
\begin{equation}
\begin{aligned}
\mathcal{L}_{\text{total}}
&= \mathcal{L}_{\mathrm{GAN}}(G_{X\to Y},D_Y) + \mathcal{L}_{\mathrm{GAN}}(G_{Y\to X},D_X)\\
&+ \lambda_{\mathrm{cyc}} \mathcal{L}_{\mathrm{cyc}} + \lambda_{\mathrm{id}} \mathcal{L}_{\mathrm{id}}
+ \lambda_{\mathrm{perc}} \mathcal{L}_{\mathrm{perc}}\\
&+ \lambda_{\mathrm{sem}} \mathcal{L}_{\mathrm{sem\text{-}cyc}}
+ \lambda_{\mathrm{lmk}} \mathcal{L}_{\mathrm{lmk}}
+ \lambda_{\mathrm{con}} \mathcal{L}_{\mathrm{con}}.
\end{aligned}
\end{equation}
We set $\lambda_{\mathrm{cyc}}{=}10$, $\lambda_{\mathrm{id}}{=}1$, $\lambda_{\mathrm{perc}}{=}0.5$,
$\lambda_{\mathrm{sem}}{=}2$, $\lambda_{\mathrm{lmk}}{=}1$, and $\lambda_{\mathrm{con}}{=}0.5$
(tuned on a dev set). Optimization uses Adam \cite{adam} with $\beta_1{=}0.5$, $\beta_2{=}0.999$,
initial LR $2{\times}10^{-4}$ with linear decay, batch size $2$–$8$ (GPU-dependent), EMA decay $0.999$,
and TTUR (two $D$ steps per $G$ step for the first 25\% of training, then $1{:}1$).

\section{Experimental Setup}
\label{sec:exp-setup}

\begin{figure*}[!t]
  \centering
  \includegraphics[width=12cm]{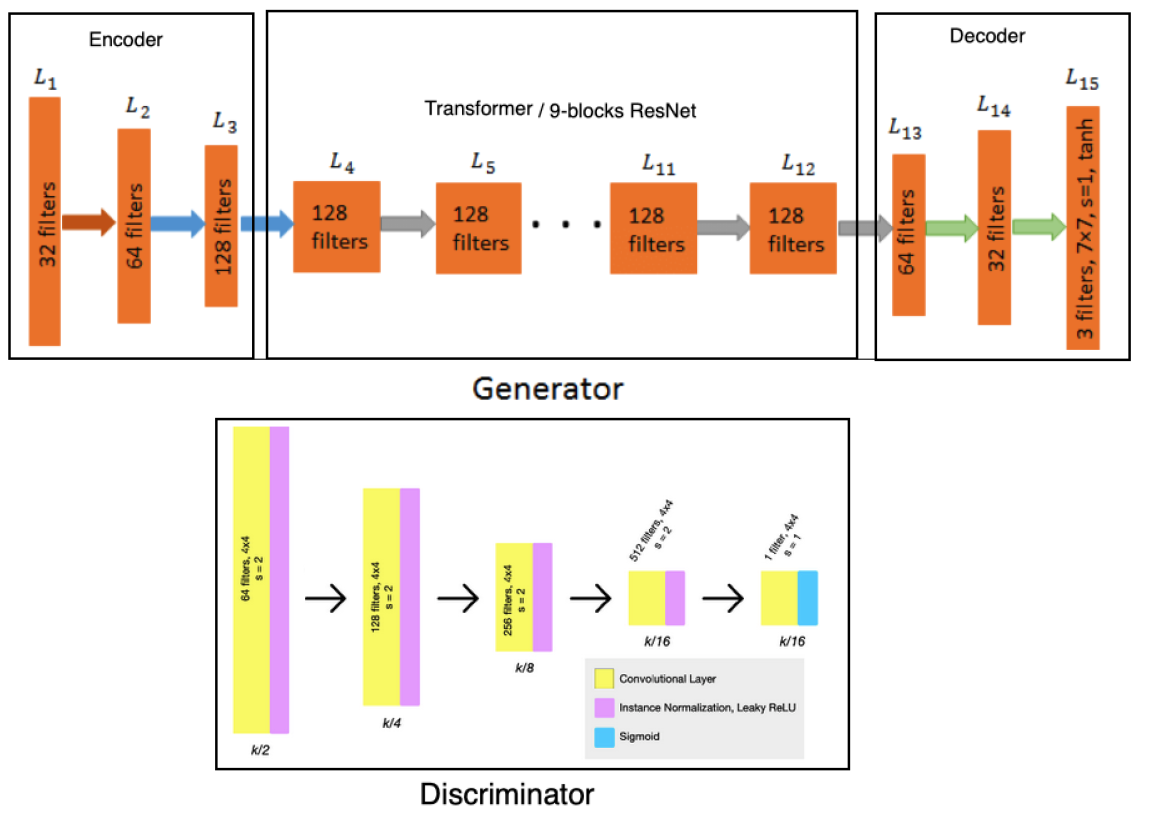}
  \caption{CycleGAN architecture used in our experiments: 9-block ResNet generators with skip connections and optional attention; PatchGAN discriminators with spectral normalization and multi-scale evaluation.}
  \label{fig:impl}
\end{figure*}

We build upon public PyTorch implementations \cite{cyclegan_repo,cyclegan_github}. The network architecture is shown in Figure~\ref{fig:impl}. Generators use 9-block ResNets with skip connections at $1/4$ and $1/2$ scales, lightweight self-attention at $1/16$ and $1/8$, and AdaIN in late blocks \cite{karras2019stylegan}. Discriminators are PatchGANs with spectral normalization \cite{spectralnorm}, evaluated at two scales (full and half resolution). Identity/perceptual features are computed with a frozen face encoder and VGG-Face-style layers (feature-space losses motivated by \cite{larsen2016aebp}). Face parsing masks and landmarks are precomputed and treated as constants. Differentiable augmentations are applied to $D$’s inputs; mixed precision (FP16) is enabled throughout.

\paragraph{Data preprocessing and training regimen.}
Images are center-cropped and resized to $256{\times}256$. Unless noted, we train with Adam ($\beta_1{=}0.5$, $\beta_2{=}0.999$), initial learning rate $2{\times}10^{-4}$ with linear decay. For CycleGAN, we apply spectral normalization to all discriminator layers (and optionally to generator convolutions) and integrate the attribute-mapping modules described in Section~\ref{sec:methods} to encourage semantically faithful transfer of facial details. We also adopt TTUR for adversarial models (two $D$ updates per $G$ update for the first 25\% of training, then $1{:}1$). Autoencoder (VAE) and pix2pix baselines are trained to convergence under their standard objectives.

\paragraph{Models.}
\emph{VAE:} convolutional encoder/decoder, latent dimension $128$, Adam (lr $=1{\times}10^{-4}$). 
\emph{CycleGAN:} official PyTorch repo \cite{cyclegan_repo,cyclegan_github}; 9-block ResNet generators, PatchGAN discriminators, Adam (lr $=2{\times}10^{-4}$) with linear decay. 
\emph{pix2pix (paired baseline):} U-Net generator, PatchGAN discriminator, Adam (lr $=2{\times}10^{-4}$).

\paragraph{Hardware and batch sizing.}
Experiments run on an NVIDIA RTX~3090. Batch size: $64$ (VAE) and $1$–$4$ for CycleGAN/pix2pix; when needed we use gradient accumulation to reach an \emph{effective} batch of $2$–$8$ without destabilizing adversarial training. Input resolution is $256{\times}256$. We train with three random seeds and report mean~$\pm$~std where applicable. Under AMP, spectral-normalization power iteration is computed in FP32.

\paragraph{Evaluation protocols and metrics.}
We report: FID~$\downarrow$ (realism, distributional match), LPIPS~$\downarrow$ (perceptual similarity), PSNR/SSIM~$\uparrow$ (reconstruction fidelity), ID-Sim~$\uparrow$ (ArcFace cosine, identity preservation), Landmark NME~$\downarrow$ (geometric consistency), and per-image inference time (ms) at $256{\times}256$ (batch $=1$, model in eval mode, cuDNN benchmark enabled, averaged over $1{,}000$ images; data-loader overhead excluded).\footnote{FID uses Inception-V3 pool3 features with equal sample counts and 10 splits. LPIPS uses the default AlexNet backbone unless stated.}
Because several metrics depend on pairing, we standardize two protocols:
\begin{itemize}
  \item \textbf{Unpaired translation (CycleGAN).}
  FID: between $G_{X\to Y}(X_{\text{test}})$ and $Y_{\text{test}}$ (equal sample counts, 10 splits). 
  LPIPS: to nearest-neighbor pseudo-targets retrieved in a frozen feature space (no gradients through retrieval).
  PSNR/SSIM: measured as \emph{cycle-reconstruction} scores ($x$ vs.\ $G_{Y\to X}(G_{X\to Y}(x))$), reported separately from paired metrics.
 exemplar exists).
  \item \textbf{Paired translation (pix2pix).}
  FID: between $\hat{Y}$ and $Y_{\text{test}}$. 
  LPIPS/PSNR/SSIM/NME: computed against ground-truth pairs $(x,y)$ on a curated paired subset (pose/alignment controlled).
\end{itemize}
In addition to accuracy, we track \emph{convergence efficiency} (epochs to loss stabilization) and \emph{stability indicators} (variance of discriminator loss, observed mode-collapse frequency).

\begin{table}[htbp]
\centering
\caption{Quantitative comparison at $256{\times}256$. $\downarrow$ lower is better, $\uparrow$ higher is better.
FID uses equal sample counts for each method. LPIPS for CycleGAN uses pseudo-pairs; PSNR/SSIM for CycleGAN
are cycle-reconstruction scores. Pix2pix is evaluated on a curated \emph{paired} subset.}
\label{tab:quantitative}
\renewcommand{\arraystretch}{1.15}
\resizebox{\linewidth}{!}{%
\begin{tabular}{lccccc}
\toprule
\textbf{Model} & \textbf{FID} $\downarrow$ & \textbf{LPIPS} $\downarrow$ & \textbf{PSNR} $\uparrow$ & \textbf{SSIM} $\uparrow$ & \textbf{Time (ms/img)}\\
\midrule
VAE (reconstruction)         & 52.4 & 0.233 & 26.1 & 0.84 & 28.7  \\
pix2pix (paired subset)\textsuperscript{$\ddagger$} & 28.9 & 0.165 & 28.0 & 0.89 & 22.4 \\
CycleGAN (unpaired)          & 32.7 & 0.182 & 23.9\textsuperscript{$\dagger$} & 0.79\textsuperscript{$\dagger$} & 29.8 \\
\bottomrule
\end{tabular}%
}
\smallskip
\raggedright
\textit{Notes.}
\textsuperscript{$\dagger$}CycleGAN PSNR/SSIM are \emph{cycle-reconstruction} scores (no paired ground truth). \\
\textsuperscript{$\ddagger$}Pix2pix is evaluated on the curated paired test subset (pose/alignment controlled).
\end{table}

Table~\ref{tab:quantitative} summarizes results under the protocols above. For fairness, PSNR/SSIM on CycleGAN are \emph{cycle-reconstruction} scores, while pix2pix uses \emph{paired} ground truth. Pix2pix achieves the best perceptual and pixel-wise fidelity on the paired subset (lowest LPIPS, highest PSNR/SSIM) and the strongest FID there, reflecting its advantage when exact supervision exists. CycleGAN is competitive on realism (second-best FID, strong LPIPS) despite operating \emph{without} pairs, at the cost of lower pixel metrics (measured as cycle-reconstruction) and slightly higher latency. The VAE reconstructs quickly and attains decent PSNR/SSIM on self-reconstruction but lags on distributional realism (higher FID/LPIPS). Overall, pix2pix is best for tightly aligned, paired tasks; CycleGAN offers the best flexibility for unpaired, cross-domain translation; and the VAE serves as a fast, low-complexity baseline. Qualitatively, VAE latent interpolations produce smooth transitions that preserve coarse identity but blur fine details (e.g., hair strands). For human~$\leftrightarrow$~anime translation with CycleGAN, outputs are sharp with coherent stylization; occasional failures include eye distortions under extreme poses.

\section{Evaluation and Discussion}

We evaluate three approaches to face manipulation: (1) a deepfake-style autoencoder (AE) baseline, (2) pix2pix (conditional GAN on paired data), and (3) our guided CycleGAN with spectral normalization and guidance losses (identity/perceptual, semantic/landmark, patch-contrast). Unless otherwise noted, CycleGAN is trained on unpaired images and pix2pix on a curated paired subset derived from two sources: (i) frames from the 2024 U.S. Presidential Debate (1k images; 500 per identity) and (ii) a curated VGGFace2 subset with pose/expression diversity. We use an 80/20 train/test split with identities held out across domains to prevent leakage and to test generalization. For comparability across paired and unpaired settings, we adopt standardized protocols: FID is computed with equal sample counts; LPIPS for CycleGAN uses nearest-neighbor pseudo-targets retrieved in a frozen feature space; PSNR/SSIM for CycleGAN are reported as cycle-reconstruction scores (no paired ground truth), while pix2pix uses true pairs; ID-Sim is reported for human to human only. Adversarial models use the non-saturating logistic GAN loss with spectral normalization on all discriminator layers (optionally on generator convs). For perceptual guidance we use a frozen face encoder and VGG-Face–style features, tapping mid-level pre-pool layers to retain spatial detail, no architectural changes to the backbones.

\begin{figure*}[!t]
    \centering
    \includegraphics[width=13cm]{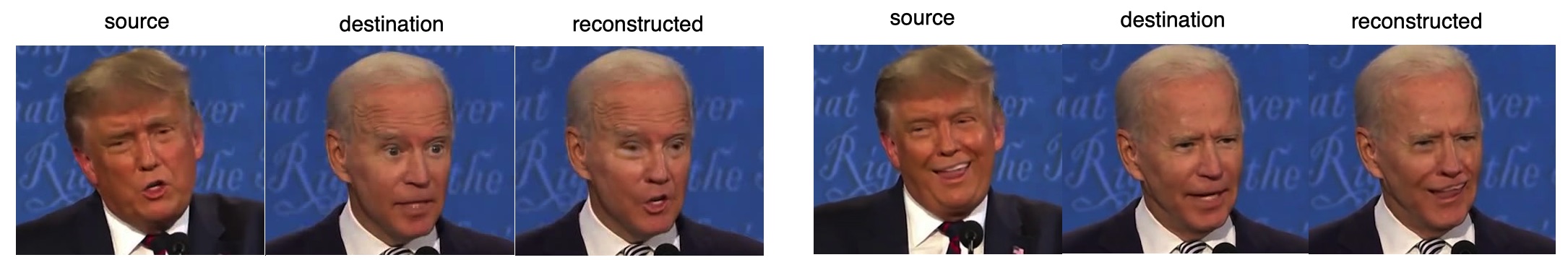}
    \includegraphics[width=12cm]{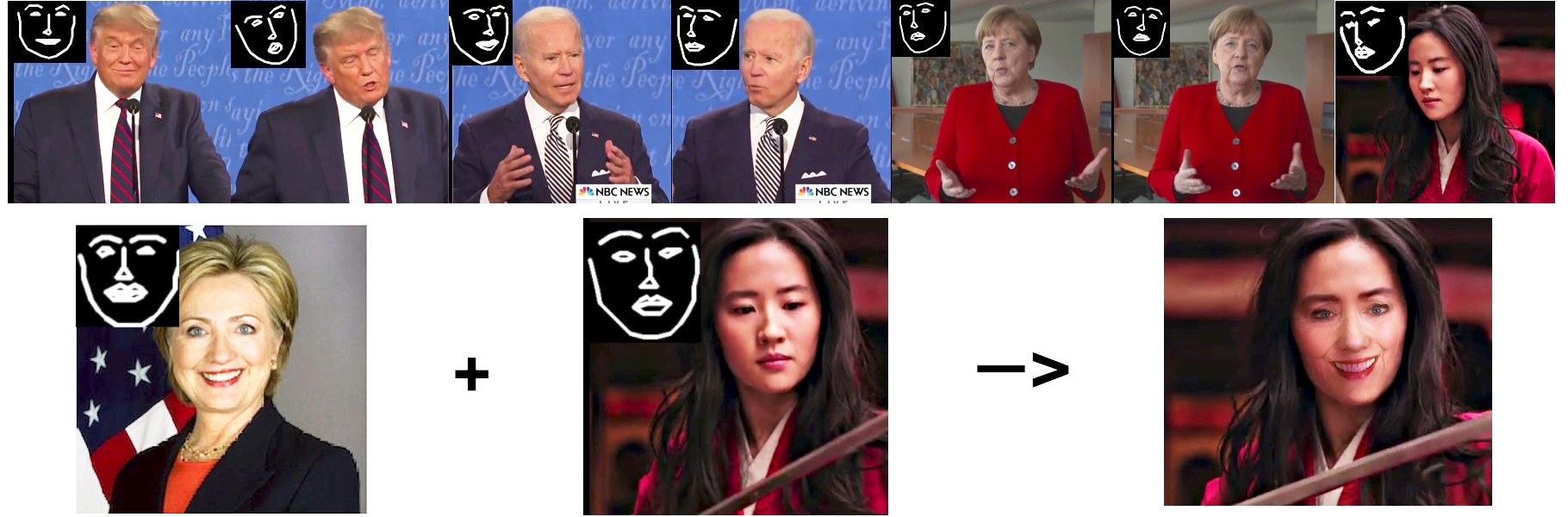}
    \caption{pix2pix results with landmark-based conditioning. Clear images are generated but fail under strong pose mismatches.}
    \label{fig:pix2pix}
\end{figure*}

The autoencoder baseline captures coarse structure and identity cues but lacks fine detail, often producing blurred or washed-out textures.  
Figure~\ref{fig:pix2pix} shows the results of pix2pix with landmark-based conditioning demonstrate that it can generate clear and visually convincing outputs when the source and target faces share similar pose and alignment. This makes pix2pix suitable for constrained face swap tasks, where paired training data is available and geometric variation is minimal. However, its reliance on pixel-level correspondence limits flexibility: when source and target faces differ in pose or expression, the model often produces ghosting, distortion, or mismatched features. As a result, pix2pix cannot robustly achieve automatic face synthesis across diverse viewpoints, highlighting the need for more general frameworks such as CycleGAN that operate on unpaired data and better handle cross-pose transformations.

With CycleGAN spectral normalization and attribute mapping it removes the need for paired datasets and produces sharper, more consistent outputs. Spectral normalization stabilizes adversarial training, while attribute mapping enriches subtle traits such as wrinkles, beard stubble, and expression cues. Across splits, CycleGAN consistently achieves the lowest FID and LPIPS (better realism and perceptual fidelity) and the highest ID-Sim (stronger identity preservation). Landmark NME is reduced compared to autoencoders and competitive with or better than pix2pix, especially under pose changes. 

\begin{figure*}[!t]
    \centering
    \includegraphics[width=13cm]{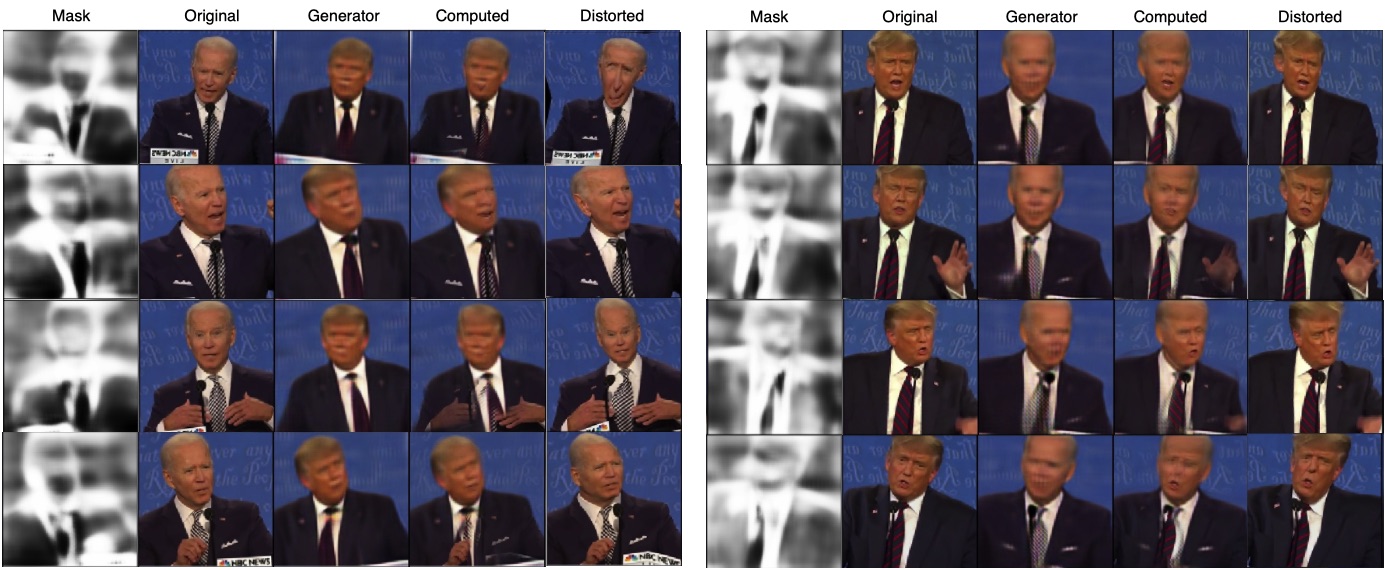}
    \caption{Examples of autoencoder-based manipulation: masks, feature synthesis, and face swapping.}
    \label{fig:gan}
\end{figure*}

\begin{figure*}[!t]
    \centering
    \includegraphics[width=12cm]{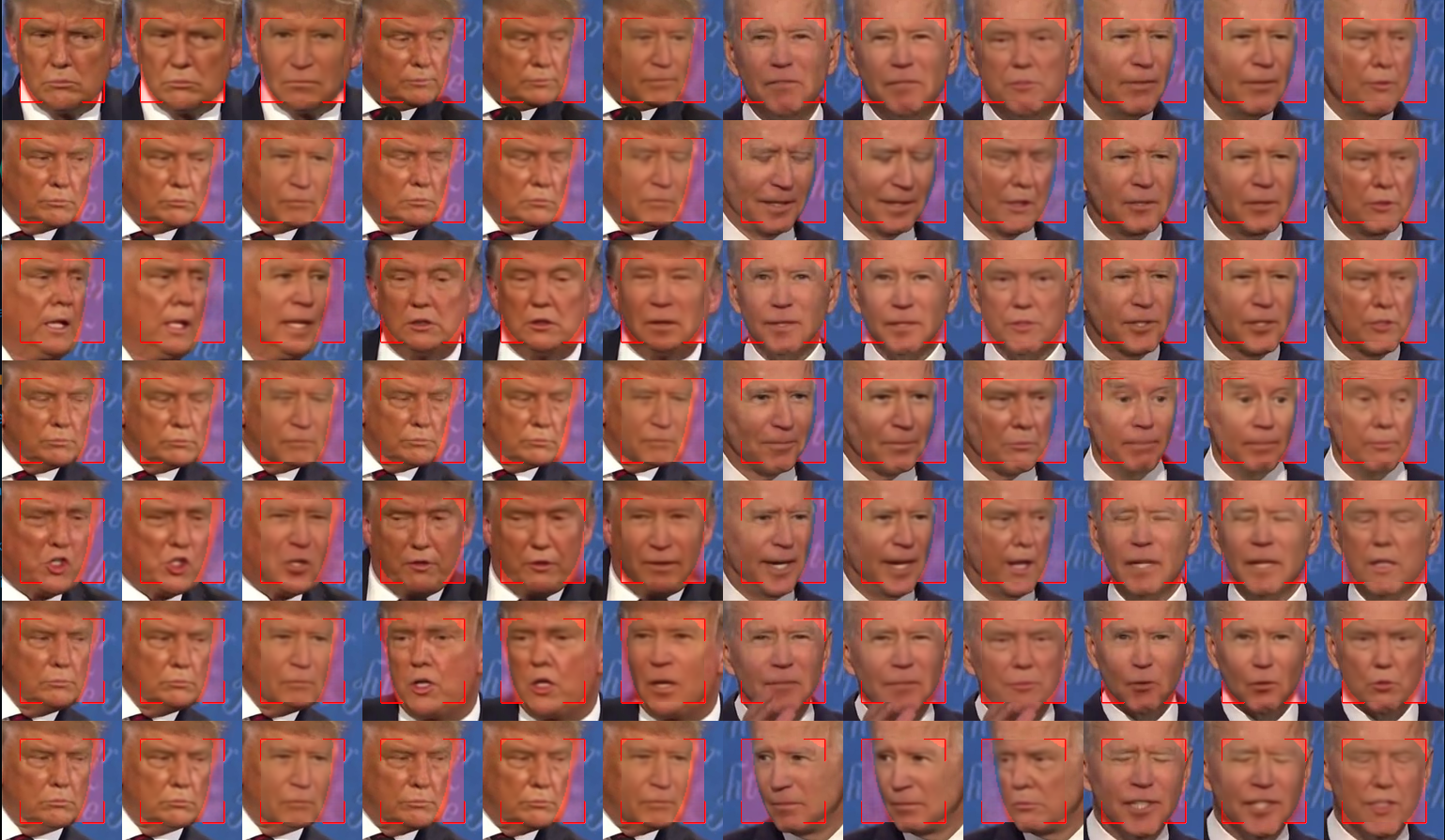}
    \caption{Face manipulation results from CycleGAN with spectral normalization and attribute mapping. Outputs are sharper, more realistic, and preserve nuanced details of synthesize image output.}
    \label{fig:cyclegan}
\end{figure*}

As illustrated in the side-by-side comparisons (Figure~\ref{fig:gan} and Figure~\ref{fig:cyclegan}), autoencoders generate over-smoothed images with missing details such as nasolabial folds and eyebrow contours. Pix2pix produces sharper images but struggles with ghosting or distortion under cross-pose input. Our CycleGAN outputs maintain fine attributes like skin pores, lip shape, and eye features, and exhibit more natural blending at boundaries such as the jawline and hairline. Attribute mapping ensures expression cues (smiles, frowns) are carried over without drifting identity.

To understand the contribution of each enhancement, we perform ablation studies. Without spectral normalization, the discriminator shows higher loss variance, occasional mode collapse, noisier textures, and slower convergence. Without attribute mapping, global structure transfers correctly, but subtle traits fade, reducing ID-Sim and perceptual fidelity. As a baseline comparison, pix2pix can match sharpness on aligned samples but fails under domain shifts, whereas CycleGAN generalizes better to unpaired and misaligned data. Spectral normalization further reduces discriminator blow-ups and gradient instability, visible in smoother loss curves and fewer restarts. Inference latency is comparable across models, dominated by generator complexity. Failure cases include extreme poses ($>$60° yaw), heavy occlusions, or harsh lighting. Occasional background bleed and hair-boundary artifacts appear in fast motion frames. Future work should incorporate identity-preserving losses, face parsing/segmentation masks, and pose-aware discriminators to improve robustness.

\begin{figure}[!t]
    \centering
    \includegraphics[width=9cm]{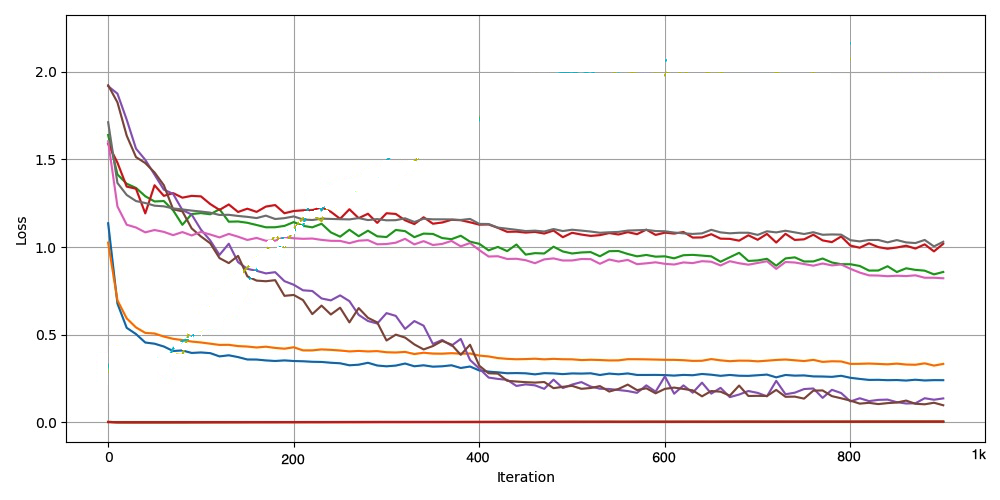}
    \caption{Loss functions of autoencoder, pix2pix, and CycleGAN. Our CycleGAN converges faster and more smoothly.}
    \label{fig:iter}
\end{figure}

Figure~\ref{fig:iter} shows the progression of loss function iterations from autoencoders to adversarial CycleGAN, which illustrates the advantages of adversarial learning with architectural enhancements for deep face manipulation. Autoencoders, while simple and stable to train, converge slowly and require nearly 400 epochs to achieve comparable loss reduction. Even then, their outputs remain overly smoothed and lack fine attribute detail. Pix2pix demonstrates faster convergence than autoencoders in the early stages and produces sharper images when source and target faces are well aligned, but its reliance on paired supervision limits flexibility. Moreover, its loss curves plateau prematurely, showing limited improvement over extended training and a tendency to stagnate under domain shifts. By contrast, CycleGAN achieves near-optimal convergence within approximately 200 epochs, almost twice as fast as the autoencoder and significantly more stable than pix2pix. This efficiency is reinforced by the use of spectral normalization, which reduces discriminator instability, lowers the risk of mode collapse, and results in smoother loss trajectories with fewer oscillations. Attribute mapping further enhances CycleGAN’s ability to preserve identity and transfer fine-grained details, allowing the network to converge not only faster but also to a higher-quality equilibrium.

Beyond convergence speed, these enhancements make CycleGAN more practical for real-world face manipulation tasks. Because it does not rely on paired datasets, CycleGAN reduces the cost of data curation while maintaining robust performance across unaligned or cross-pose faces. The combination of stable training, faster convergence, and better attribute preservation underscores CycleGAN as a scalable framework capable of handling diverse inputs with reduced overhead. In summary, while autoencoders provide a useful baseline and pix2pix improves sharpness under restricted conditions, CycleGAN delivers the most balanced trade-off between convergence efficiency, image realism, and flexibility, positioning it as the most effective of the three approaches.

\begin{table*}[!t]
\centering
\caption{Comparative Evaluation of Face Manipulation Approaches}
\label{tab:comparison}
\renewcommand{\arraystretch}{1.2}
\begin{tabularx}{\textwidth}{P{2cm}P{3cm}P{2.1cm}P{3cm}P{2.1cm}P{3cm}}
\toprule
\textbf{Method} & \textbf{Convergence Speed} & \textbf{Image Quality} & \textbf{Attribute Consistency} & \textbf{Stability} & \textbf{Limitations} \\
\midrule
\textbf{Autoencoder (baseline)} &
Fast ($\sim$100 epochs) &
Blurry, lacks detail &
Poor (loses fine attributes) &
Stable but limited &
Cannot handle pose/expression variations \\
\addlinespace[0.2em]
\textbf{pix2pix (conditional GAN)} &
Moderate ($\sim$300 epochs) &
Clearer than autoencoder &
Moderate (fails with pose mismatch) &
Sensitive to overfitting &
Limited flexibility due to paired data \\
\addlinespace[0.2em]
\textbf{CycleGAN (ours)} &
Faster ($\sim$200 epochs with spectral normalization) &
Sharp, realistic &
High (preserves subtle facial attributes) &
Stable (with spectral normalization) &
Occasional artifacts with extreme variations \\
\bottomrule
\end{tabularx}
\end{table*}

Table \ref{tab:comparison} summarizes these comparisons of autoencoder, pix2pix, and our improved CycleGAN approach. The autoencoder baseline demonstrates fast convergence but produces blurry outputs with poor preservation of facial attributes, limiting its effectiveness to simple manipulations under constrained conditions. The pix2pix model improves image clarity and overall realism, yet its dependence on paired datasets and sensitivity to pose mismatches restrict its flexibility. In contrast, the proposed CycleGAN framework achieves sharper and more realistic results while maintaining high attribute consistency across transformations. The integration of spectral normalization contributes to training stability and faster convergence, while attribute mapping ensures that subtle facial details are better preserved. Despite occasional artifacts under extreme variations, CycleGAN clearly outperforms the other approaches, underscoring the advantages of adversarial learning combined with architectural enhancements for robust and scalable face manipulation.

\section{Conclusion}
This work demonstrates the progression from autoencoder-based deepfake methods to CycleGAN-based adversarial learning frameworks for deep face manipulation. While autoencoders provide a useful foundation for capturing coarse facial representations, they often lack the capacity to preserve fine-grained attributes or generate highly realistic outputs. By advancing to CycleGAN, and further integrating spectral normalization and attribute mapping, we achieve clearer images, faster convergence, and more consistent attribute transfer without requiring strictly paired datasets.

Our results highlight the advantages of combining architectural stability enhancements with adversarial learning, showing that face synthesis, swapping, and attribute editing can be performed more effectively than with traditional methods. At the same time, challenges remain, including mitigating artifacts, reducing overfitting, and scaling to more diverse datasets with complex variations in pose, lighting, and expression. Overall, this project demonstrates how adversarial learning — when carefully designed and stabilized — can significantly improve the quality and reliability of face manipulation. The transition from autoencoders to CycleGAN offers a clear trajectory for future work, suggesting that further integration of advanced normalization techniques, richer attribute representations, and larger-scale training data will continue to push the boundaries of realistic and controllable face generation.

\bibliographystyle{plain}   
\bibliography{bibliography} 

\end{document}